\documentclass[10pt,twocolumn,letterpaper]{article}

\usepackage[pagenumbers]{iccv} 

\newcommand{\Ours}{MotionMatcher}
\newcommand{\ours}{{\Ours}}
\newcommand{\gt}{\hat{v_t}}
\newcommand{\pd}{v^{\theta}_t}
\newcommand{\E}{\mathbb{E}_{z_0, t, \epsilon}}
\newcommand{\mca}{M_{{\rm CA}}}
\newcommand{\mtsa}{M_{{\rm TSA}}}
\newcommand{\wca}{\lambda_{{\rm CA}}}
\newcommand{\wtsa}{\lambda_{{\rm TSA}}}
\newcommand{\mfe}{{\mathcal M}}
\DeclareMathOperator*{\argmin}{arg\,min}

\definecolor{iccvblue}{rgb}{0.21,0.49,0.74}
\usepackage[pagebackref,breaklinks,colorlinks,allcolors=iccvblue]{hyperref}

\usepackage{svg}
\usepackage{float}
\usepackage{makecell}

\def\paperID{*****} 
\def\confName{ICCV}
\def\confYear{2025}

\title{\Ours: Motion Customization of Text-to-Video Diffusion Models\\via Motion Feature Matching}

\author{
Yen-Siang Wu$^{1,\dagger}$, Chi-Pin Huang$^1$, Fu-En Yang$^2$, Yu-Chiang Frank Wang$^{1,2,\ddagger}$\\
$^1$National Taiwan University\\
$^2$NVIDIA\\
{\tt\small $^{\dagger}$b09902097@ntu.edu.tw, $^{\ddagger}$frankwang@nvidia.com}\\
{\tt\normalsize\href{https://www.csie.ntu.edu.tw/~b09902097/motionmatcher/}{\color{Magenta}{https://b09902097.github.io/motionmatcher/}}}
}

\begin{document}
\twocolumn[{
\maketitle
\begin{center}
    \centering
    \includegraphics[width=0.865\textwidth]{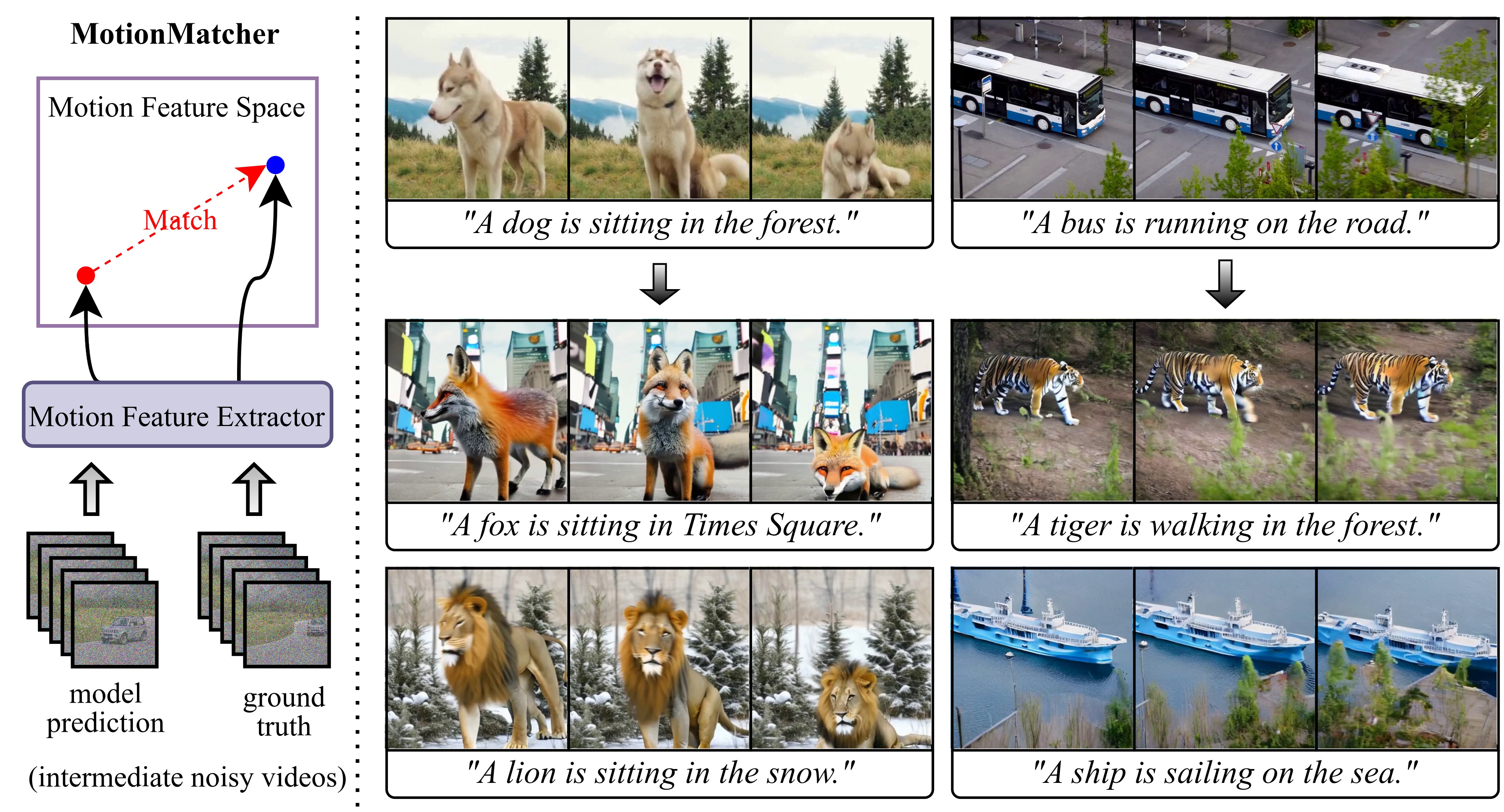}
    \captionof{figure}{{\bf\ours} can customize pre-traind T2V diffusion models with a user-provided reference video (top row). Once customized, the diffusion model is able to transfer the precise motion (including object movements and camera framing) in the reference video to a variety of scenes (middle and bottom rows).}
\label{fig:teaser}
\end{center}
}]
\begin{abstract}
Text-to-video (T2V) diffusion models have shown promising capabilities in synthesizing realistic videos from input text prompts. However, the input text description alone provides limited control over the precise objects movements and camera framing. In this work, we tackle the motion customization problem, where a reference video is provided as motion guidance. While most existing methods choose to fine-tune pre-trained diffusion models to reconstruct the frame differences of the reference video, we observe that such strategy suffer from content leakage from the reference video, and they cannot capture complex motion accurately. To address this issue, we propose {\ours}, a motion customization framework that fine-tunes the pre-trained T2V diffusion model at the feature level. Instead of using pixel-level objectives, {\ours} compares high-level, spatio-temporal motion features to fine-tune diffusion models, ensuring precise motion learning. For the sake of memory efficiency and accessibility, we utilize a pre-trained T2V diffusion model, which contains considerable prior knowledge about video motion, to compute these motion features. In our experiments, we demonstrate state-of-the-art motion customization performances, validating the design of our framework.
\end{abstract}
\section{Introduction}
\label{sec:intro}

To control the rhythm of a movie scene, movie directors would carefully arrange the precise movements and positioning of both the actors and the camera for each shot (as known as staging/blocking). Similarly, to control the pacing and flow of AI-generated videos, users should have control over the dynamics and composition of videos produced by generative models. To this end, numerous motion control methods~\cite{dragnuwa, mctrl, draga, mb, trailblazer, peekaboo, boximator} have been proposed to control moving object trajectories in videos generated by text-to-video (T2V) diffusion models~\cite{vdm, vldm}. Motion customization, in particular, aims to control T2V diffusion models with the motion of a reference video~\cite{mc, md, vmc, dmt, sma}. With the assistance of the reference video, users are able to specify the desired object movements and camera framing in detail. Formally speaking, given a reference video, motion customization aims to adjust a pre-trained T2V diffusion model, so the output videos sampled from the adjusted model follow the object movements and camera framing of the reference video (see \cref{fig:teaser} for an example). Given that motion is a high-level concept involving both spatial and temporal dimensions~\cite{moft, dmt}, motion customization is considered a non-trivial task.

Recently, many motion customization methods have been proposed to eliminate the influence of visual appearance in the reference video. Among them, a standout strategy is fine-tuning the pre-trained T2V diffusion model to reconstruct the frame differences of the reference video. For instance, VMC~\cite{vmc} and SMA~\cite{sma} use a motion distillation objective that reconstructs the residual frames of the reference video. MotionDirector~\cite{md} proposes an appearance-debiased objective that reconstructs the differences between an anchor frame and all other frames. However, we find that frame differences do not accurately represent motion. For example, two videos with the same motion, such as a red car and a blue car both driving leftward, can yield completely different frame differences because the pixel changes occur in different color channels in each video. Moreover, since frame differences only process videos at the pixel level, they cannot capture complex motion that requires a high-level understanding of video, such as rapid movements or movements in low-texture regions. In these cases, the strategy of reconstructing frame differences fails to reproduce the target motion.

To address this issue, we propose {\ours}, a novel fine-tuning framework for motion customization via motion feature matching. Instead of aligning pixel values or frame differences as in previous methods, {\ours} aligns the projected motion features extracted from a pre-trained feature extractor. Since these motion features are calculated with a sophisticated pre-trained model, they are capable of capturing complex motion that requires a high-level, spatio-temporal understanding of video. This effectively addresses the limitation of previous work, where frame differences fail to capture complex motion.

{\ours} differs from traditional fine-tuning approaches. At each fine-tuning step, it starts off by using a feature extractor to compute the motion features of the output video and the motion features of the reconstruction ground truth video. Our feature matching objective then minimizes the L2 distance between the two feature vectors. However, since the output videos of T2V diffusion models are in latent space and at certain noise levels, the feature extractor must be able to process latent noisy videos. To obtain such a feature extractor, we take advantages of (1) pre-trained T2V diffusion models' ability in extracting features from noisy, latent videos and (2) the spatio-temporal information encoded in attention maps. We find that cross-attention maps (CA) in pre-trained diffusion models contain information about camera framing, while temporal self-attention maps (TSA) represent object movements. Therefore, we utilize them to represent motion features. Ultimately, the design of our framework is validated through detailed analysis and extensive experiments.

To summarize, our key contributions include:
\begin{itemize}
  \item We propose {\bf\ours}, a feature-level fine-tuning framework for motion customization. It leverages a pre-trained feature extractor to map videos into a motion feature space, capturing high-level motion information. By aligning the motion features, the diffusion model learns to generate videos with the target motion.
  \item To extract features from \emph{noisy latent videos}, we utilize the pre-trained diffusion model as a feature extractor, as it naturally processes such inputs.
  \item We identify two sources of motion cues---cross-attention maps and temporal self-attention maps---and use them to form the motion features.
  \item We demonstrate that {\ours} achieves state-of-the-art performance through comprehensive experiments. It offers superior joint controllability of text and motion, advancing scene staging in AI-generated videos.
\end{itemize}
\section{Related work}
\label{sec:related_work}

\subsection{Text-to-video generation}
\label{sec:t2v}

Text-to-video (T2V) generation models aim to synthesize videos that comply with user-provided text descriptions. Previously, a large number of T2V models have been proposed, including GANs~\cite{gan1, gan2, gan3, gan4}, autoregressive models~\cite{ar1, ar2, ar3, ar4}, and diffusion models~\cite{vdm, vldm, cogvideo}.

Following the success of text-to-image (T2I) diffusion models~\cite{dm1, dm2, dm3}, researchers have also put considerable effort into training T2V diffusion models recently. To achieve this, a commonly used approach is inflating a pre-trained T2I diffusion model by inserting temporal layers and finetuning the whole model on video data~\cite{modelscope, videocrafter, latentvdm, imagen, makeavideo, show1, lavie}. On the other hand, models like AnimateDiff~\cite{animatediff} and VideoLDM~\cite{vldm} also insert additional temporal layers, but they only finetune the newly-added temporal layers for decoupling purposes. In contrast to the first approach, these models are typically limited to generating simple motion~\cite{animatezero}. To ensure motion complexity, we adopt the former type of model as the base model in this work.

\subsection{Motion control in T2V generation}
\label{sec:motion_control}

To enable detailed control over camera framing and object movements in T2V generation, recent research has explored trajectory-based~\cite{dragnuwa, mctrl, draga, moft}, box-based~\cite{mb, trailblazer, peekaboo, boximator}, and reference-based motion control. Trajectory-based and box-based motion control are typically achieved by conditioning T2V diffusion models on additional motion signal and training them on large video datasets~\cite{dragnuwa, draga, mctrl, boximator}, or by directly manipulating attention maps at the inference stage~\cite{mb, trailblazer, peekaboo}. However, these approaches require users to explicitly define the trajectories of moving objects within frames, which is usually laborious and provides limited control over the entire scene. In contrast, reference-based motion control can specify the target motion more comprehensively via a reference video~\cite{mc, md, vmc, dmt, sma}. In this work, we focus on motion customization, which is considered reference-based motion control.

\subsection{Motion customization of T2V diffusion models}
\label{sec:motion_customization}

Recently, motion customization has emerged as a new area of research. It adapts the pre-trained T2V diffusion model to generate videos that replicate the camera framing and object movements of a user-provided reference video. To avoid learning visual appearance, VMC~\cite{vmc} and SMA~\cite{sma} fine-tune the pre-trained T2V diffusion model by aligning the residual frames of the output video with the residual frames of the reference video. MotionDirector~\cite{md} proposes a dual-path fine-tuning method to avoid learning visual appearance and simultaneously utilizes an objective that matches frame differences. However, since frame differences do not accurately represent motion, these methods struggle to replicate complex motion.

Another strategy is using diffusion guidance~\cite{cfg, dsg, dragon} to achieve controllable generation. Specifically, DMT~\cite{dmt} employs the intermediate spatio-temporal features in diffusion models as a guidance signal, whereas MotionClone~\cite{mc} uses intermediate temporal attention maps for guidance. Despite being training-free, these methods need to compute additional gradients during inference, resulting in a lengthy sampling process. Moreover, as noted in~\cite{gw1, gw2}, the large guidance weights used in diffusion guidance can lead to the generation of out-of-distribution samples.

While other motion customization approaches exist, they address different tasks. For instance, DreamVideo~\cite{dreamvideo} and Customize-A-Video~\cite{cav} focus solely on replicating object movements without preserving the camera framing, whereas MotionMaster~\cite{mm} deals exclusively with camera movements. In contrast, our method provides control over both object movements and camera framing.
\section{Method}
\label{sec:method}

\begin{figure*}[t]
  \centering
  \includegraphics[width=1.0\linewidth]{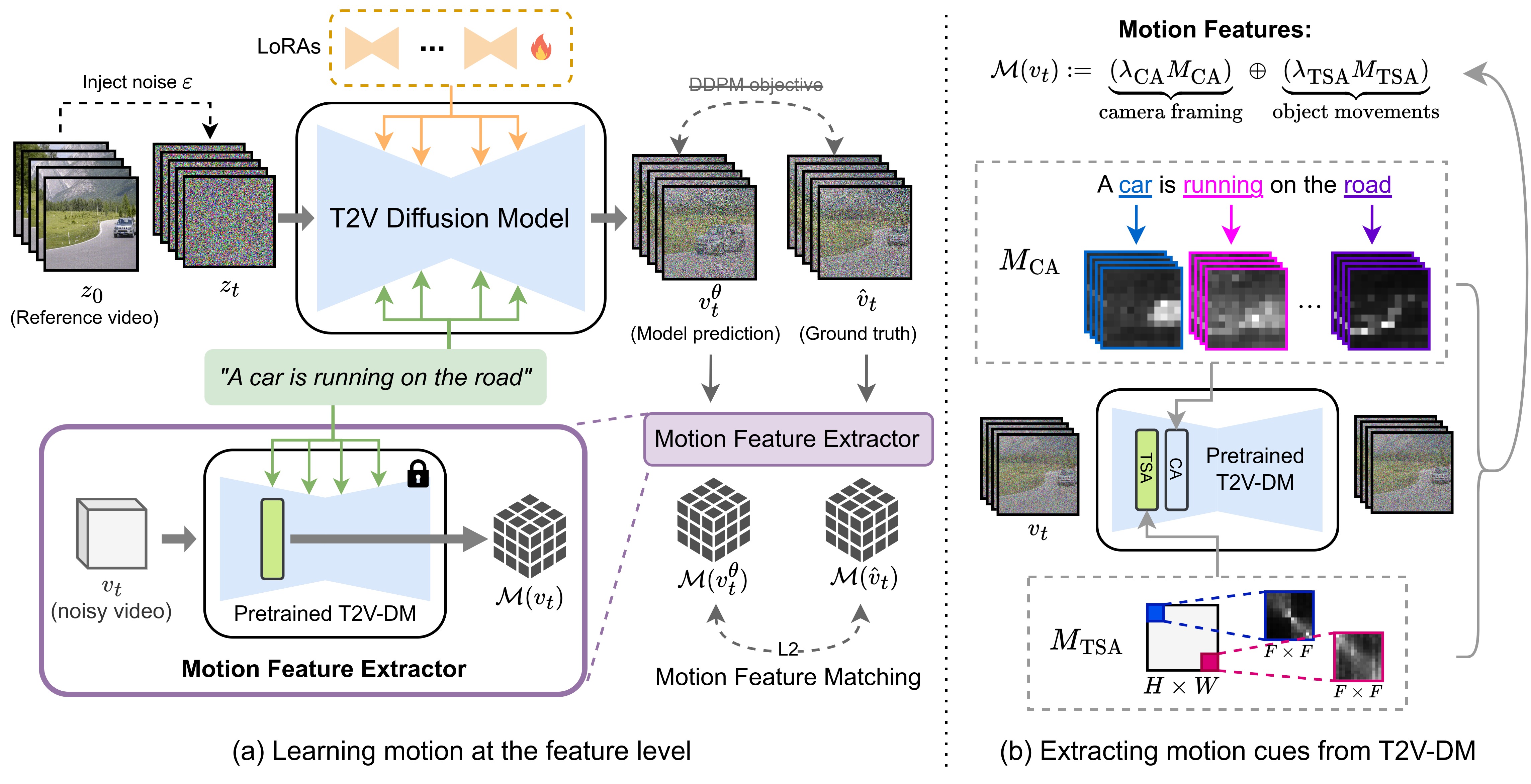}
  \caption{\textbf{Overview of {\ours}.}  (a) We fine-tune the pre-trained T2V diffusion model (T2V-DM) using the \emph{motion feature matching} objective. Unlike the standard \emph{pixel-level} DDPM loss, we align the motion features of the predicted noisy video $\pd$ with those of the ground truth noisy video $\gt$. To extract motion features from \emph{noisy latent videos}, we use a pre-trained T2V-DM (frozen) as a feature extractor. (b) We leverage the cross-attention (CA) maps and temporal self-attention (TSA) maps in the pre-trained T2V diffusion model to extract motion cues. The final motion features are the combination of the CA maps and TSA maps.}
  \label{fig:diagram}
\end{figure*}

\paragraph{Problem formulation}
To control scene staging in AI-generated videos, we tackle the problem of motion customization, specifically as defined in DMT~\cite{dmt}. Given a reference video $z_0$ and a text prompt $y$ associated with it, we aim to adjust a pre-trained T2V diffusion model $\epsilon_{\theta}$, so that the output videos sampled from the adjusted model replicate both the \emph{object movements} and \emph{camera framing} in $z_0$.

\subsection{Preliminary: Text-to-video diffusion models}

Text-to-video (T2V) diffusion models are probabilistic generative models that synthesize videos by gradually denoising a sequence of randomly sampled Gaussian noise frames (in latent space), guided by a textual condition $y$.

\paragraph{Architecture}
To model temporal information, T2V diffusion models typically inflate a pre-trained text-to-image (T2I) diffusion model by inserting temporal layers. These temporal layers are made up of feedforward networks and temporal self-attentions, where \emph{temporal self-attentions} (TSA) apply self-attention along the frame axis.

\paragraph{Training}
T2V diffusion models $\epsilon_{\theta}$ are trained by minimizing a weighted noise-prediction objective:
\begin{equation}
    \E\left[w_t\left\|
\epsilon-\epsilon_{\theta}(z_t, t, y)
\right\|^2\right]
  \label{eq:ddpm_eps},
\end{equation}
where $z_t=\sqrt{\bar\alpha_t}z_0+\sqrt{1-\bar\alpha_t}\epsilon$ is the noised video at timestep $t$, $\epsilon\sim\mathcal N(\bf{0},\bf{I})$ is Gaussian noise, and $w_t$ is a time-dependent weighting term. This noise-prediction objective is also equivalent to predicting the previous noised video at timestep $t-1$ through a different parametrization~\cite{ddpm}:
\begin{equation}
    \E\left[w'_t\left\|
v_t(z_t,\epsilon)-v_t(z_t,\epsilon_{\theta}(z_t, t, y))
\right\|^2\right]
  \label{eq:ddpm_v},
\end{equation}
where $v_t(z_t,\epsilon):=\frac 1 {\sqrt{\alpha_{t}}}z_t+\left(-\frac {\sqrt{1-\bar\alpha_t}} {\sqrt{\alpha_{t}}}+\sqrt{1-\bar\alpha_{t-1}}\right)\epsilon$ is a function that estimates the previous noised video $z_{t-1}$ based on the current video state $z_t$ and noise $\epsilon$, and $w'_t$ is the time-dependent weight after reparametrization (See supplementary material for more details). For simplicity, we will use $\pd$ to denote the model prediction $v_t(z_t,\epsilon_{\theta}(z_t, t, y))$, and use $\gt$ to denote the ground truth $v_t(z_t,\epsilon)$. The objective can therefore be rewritten as:
\begin{equation}
    \E\left[w'_t\left\|
\gt-\pd
\right\|^2\right]
  \label{eq:ddpm_sv},
\end{equation}
where $w'_t$ is the time-dependent weight in \cref{eq:ddpm_v}.

\subsection{Learning motion at the feature level}

Identifying motion in video requires a \emph{high-level} understanding of both the spatial and temporal aspects of the video, so using the standard \emph{pixel-level} DDPM reconstruction loss (\cref{eq:ddpm_sv}) for motion customization cannot accurately learn motion, and may introduce irrelevant information, such as content and visual appearance.

To this end, we introduce the \emph{motion feature matching} objective, where a deep feature extractor $\mfe$ is used to extract motion information from videos at a high level. Instead of directly aligning the predicted noisy video $\pd$ with the ground truth $\gt$ at the pixel level, we align their high-level motion features (extracted by $\mfe$):
\begin{equation}
    \mathcal{L}_{\rm mot}(\theta)=\E\left[w'_t\left\|
\mfe(\gt)-\mfe(\pd)
\right\|^2\right],
  \label{eq:mfm}
\end{equation}
where $\mfe$ is a motion feature extractor for \emph{noisy latent videos}, and $w'_t$ is the time-dependent weight in \cref{eq:ddpm_sv}. As illustrated in \cref{fig:diagram}(a), this \emph{motion feature matching} objective aims to minimize the L2 discrepancy between the two videos in the motion feature space, ensuring that the motion in output video matches the motion in the reference video.

However, designing the motion feature extractor $\mfe$ in \cref{eq:mfm} is non-trivial, as it needs to extract features from \emph{noisy latent videos}. First of all, most feature extractors, such as ViViT~\cite{vivit}, EfficientNet~\cite{eff}, DenseNet-201~\cite{densenet}, and ResNet-50~\cite{resnet}, are trained on clean visual data, so we cannot directly applied them to noisy videos. Secondly, since the videos $\gt$ and $\pd$ in \cref{eq:mfm} are in latent space, our feature extractor must be designed to process \emph{latent videos} directly. Otherwise, we would need to decode them back into pixel-space videos before applying off-the-shelf feature extractors. This would incur substantial computational and memory overhead during training, due to both backpropagation through the large VAE decoder and the cost of processing ``full-resolution'' videos. 

Here we claim that the pre-trained T2V diffusion model serve as a proper feature extractor for \emph{noisy latent videos}. Firstly, recent work has shown both theoretically and experimentally that pre-trained diffusion models are capable of extracting high-level semantics and structural information from visual data, making them a ``unified feature extractor''~\cite{ins1, ins2}. Secondly, since diffusion models are trained on \emph{noisy latent inputs}, using them as feature extractors for \emph{noisy latent videos} helps prevent a training-inference gap. For these reasons, {\ours} leverages the \textbf{pre-trained T2V diffusion model} as the motion feature extractor $\mfe$.

\subsection{Extracting motion cues from diffusion models}

In this section, we identify the locations within the intermediate layers of diffusion models from which motion-specific features can be extracted.

\paragraph{Extracting cues for camera framing}
Recent studies have shown that the cross-attention (CA) maps in diffusion models closely reflect the spatial arrangement of objects within the frame~\cite{boxdiff, db, trailblazer, peekaboo, direct}. Building on this, we leverage the CA maps from T2V diffusion models to describe the composition of each video frame (see \cref{fig:diagram}(b)), thereby determining the camera framing throughout the video (\eg, shot size and composition).

\begin{figure}[t]
  \centering
  \includegraphics[width=1.0\columnwidth]{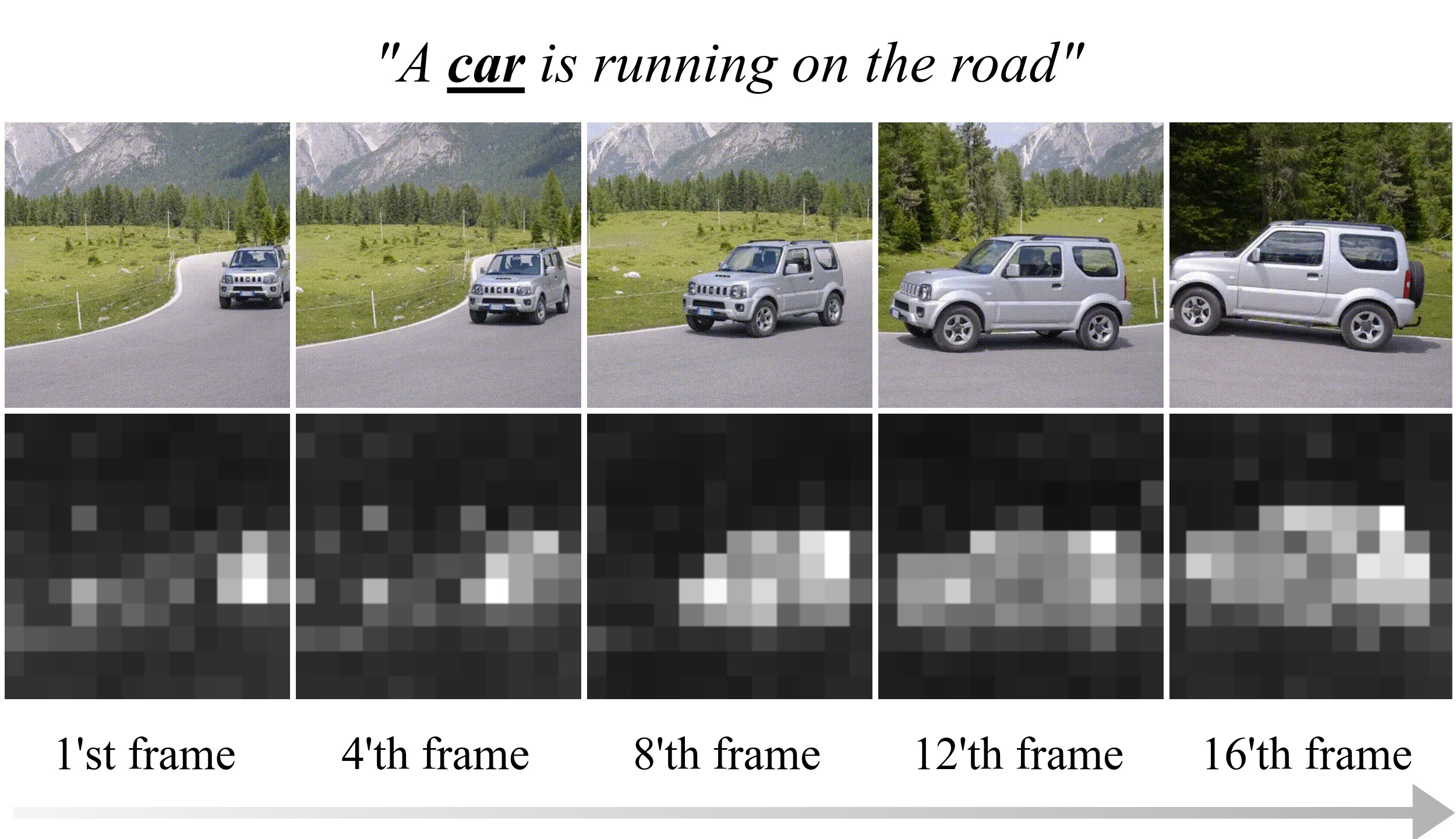}
  \caption{{\bf Example of cross-attention maps.} We visualize the cross-attention map $\mca$, computed between the activations in T2V diffusion models and the text prompt $y$. Here we obtain the CA map by adding noise to the video and using the pre-trained diffusion model as a feature extractor. The extracted CA maps reveal the placement and shot sizes of the object associated with the word ``car'' in each video frame.}
  \label{fig:mca}
\end{figure}

Formally speaking, CA maps are calculated by first reshaping the intermediate 3D activations $\Phi\in\mathbb{R}^{H\times W\times F\times D}$ into the shape $(H\times W\times F)\times D$, where $F$, $H$, $W$, and $D$ denote the number of frames, height, width, and depth of the activations. Cross-attention is then performed between the activations $\Phi$ and word embeddings $\tau(y)$ as follows :
\begin{equation}
\mca=\mathrm{Softmax}\left(\frac{Q(\Phi)K(\tau(y))^T}{\sqrt{D}}\right)
\label{eq:features},
\end{equation}
where $\tau$ denotes the text encoder used in the T2V diffusion model, and $y$ is the text prompt given by the user. In $\mca\in\mathbb[0,1]^{F\times H\times W \times |c|}$, each element $(\mca)_{i,j,k,l}$ represents the correlation between the spatial-temporal coordinate $(i,j,k)$ and the $l$'th word in the text prompt. As shown in \cref{fig:mca}, $\mca$ highlights the region within the frame that corresponds to an object. It focuses on structural information and eliminates visual appearance.

\paragraph{Extracting cues for object movements}
Since cross-attention maps cannot describe motion that does not involve spatial shifts (\eg, rotation and non-rigid motion), it is crucial to extract additional cues to represent such object movements. Since we discover that the temporal self-attention (TSA) maps in T2V diffusion models can capture detailed object movements, we also incorporate them into the motion features (see \cref{fig:diagram}(b)).

\begin{figure}[t]
  \centering
  \includegraphics[width=1.0\columnwidth]{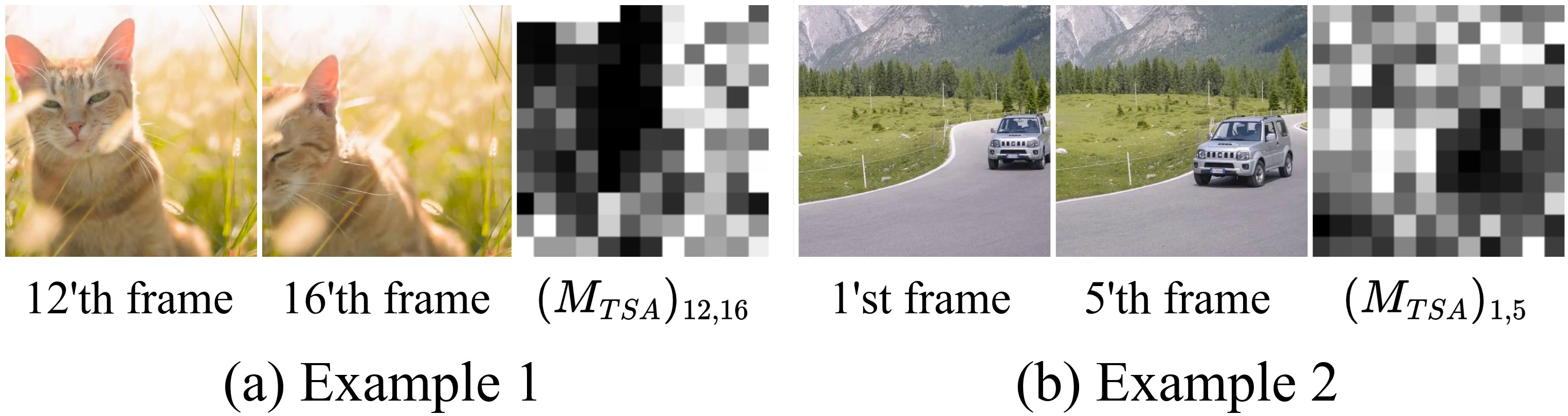}
  \caption{{\bf Example of temporal self-attention maps.} We visualize the temporal self-attention map $\mca$, computed between two different frames. Here we obtain the TSA map by adding noise to the video and using the pre-trained diffusion model as a feature extractor. The extracted TSA maps describe the dynamics of the video in detail.}
  \label{fig:mtsa}
\end{figure}

To compute temporal self-attention (TSA) maps $\mtsa$, we begin by reshaping the model's intermediate 3D activations $\Phi\in\mathbb{R}^{H\times W\times F\times D}$ into the shape $(H\times W)\times F\times D$. For each particular spatial coordinate $(i,j)$, we compute the self-attention weights between frames as follows:
\begin{equation}
(\mtsa)_{i,j}=\mathrm{Softmax}\left(\frac{Q(\Phi_{i,j})K(\Phi_{i,j})^T}{\sqrt{D}}\right)
\label{eq:features},
\end{equation}
where $i$ and $j$ denote the spatial coordinates. Specifically, each element $(\mtsa)_{i,j,k,l}$ of the TSA map $\mtsa\in [0,1]^{H\times W\times F\times F}$ represents the degree of relevance between the $k$'th and $l$'th frames at the spatial coordinate $(i,j)$, capturing the dynamics of the video. As visualized in ~\cref{fig:mtsa}, the darker regions, which indicate low correlation between frames, correspond closely to areas where significant changes occur between the two frames. Therefore, by collecting the TSA maps for all $F\times F$ frame pairs, we can capture the inter-frame dynamics in detail.

With the cross-attention maps capturing camera framing, and the temporal self-attention maps reflecting object movements, we combine both to form the motion features:
\begin{equation}(\wca\mca)\oplus (\wtsa\mtsa)
\label{eq:features},
\end{equation}
where $\wca$ and $\wtsa$ are weights that control the contributions of each component.

\begin{figure*}[htp]
  \centering
  \includegraphics[width=1.0\linewidth]{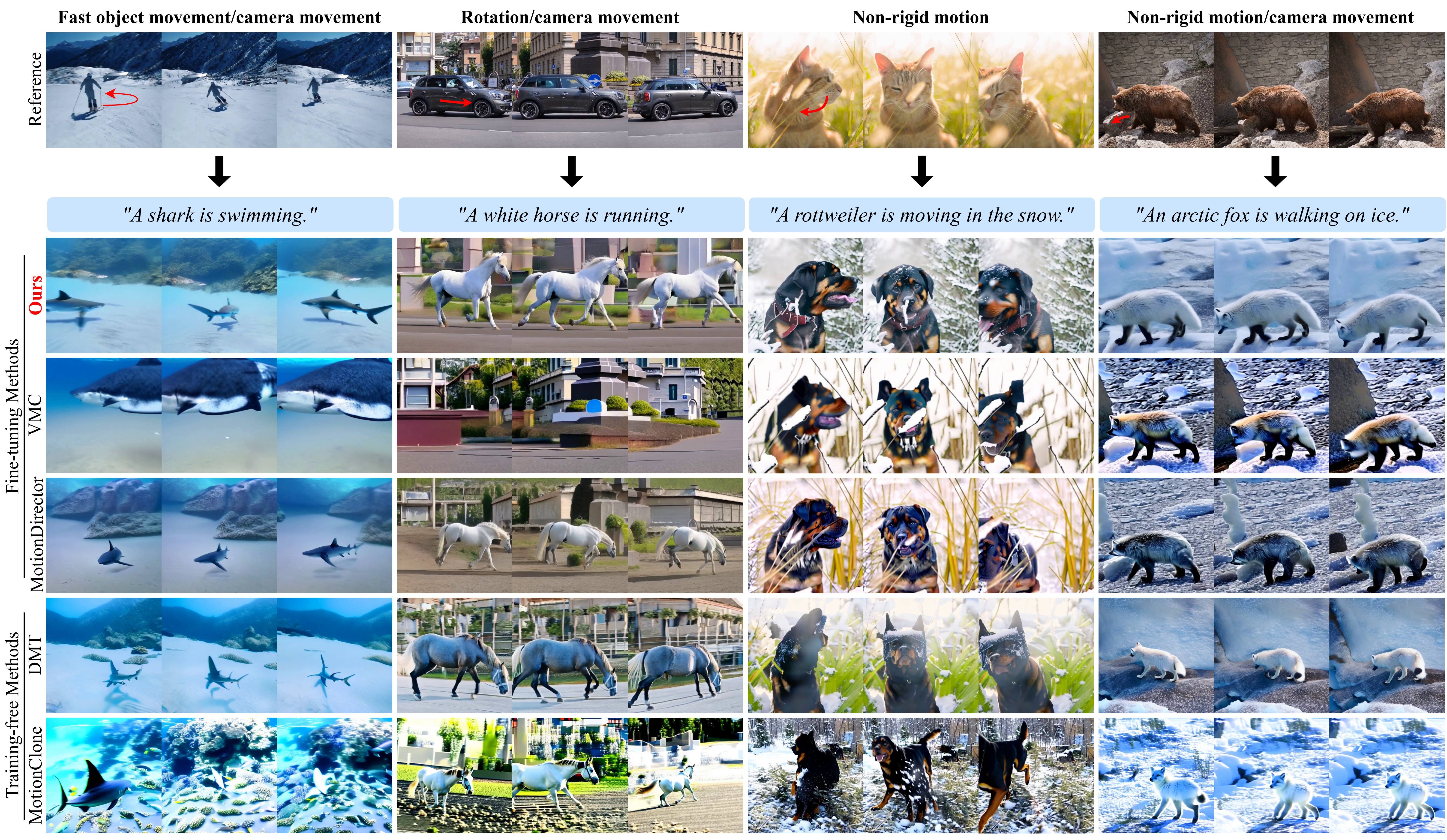}
  \caption{\textbf{Qualitative comparisons.} Compared to existing methods such as VMC~\cite{vmc}, MotionDirector~\cite{md}, DMT~\cite{dmt}, and MotionClone~\cite{mc}, our approach demonstrates superior text alignment and video quality, achieving high-fidelity motion transfer from reference videos to new scenes.}
  \label{fig:qualitative}
\end{figure*}

\subsection{Motion-aware LoRA fine-tuning}

After extracting the motion features, we fine-tune the pre-trained T2V diffusion model using the \emph{motion feature matching} objective in \cref{eq:mfm}. By aligning the $\mca$ component, we ensure that the \emph{camera framing} in the generated video matches that of the reference video, and aligning $\mtsa$ ensures that the \emph{dynamics} in the generated video align with those of the reference video.

To preserve the model's pre-trained knowledge while fine-tuning, we apply low-rank adaptations (LoRAs)~\cite{lora} to fine-tune the model with fewer trainable parameters: 
\begin{equation}
    \argmin_{\Delta\theta} \mathcal{L}_{\rm mot}(\theta+\Delta\theta),
\end{equation}
where $\Delta\theta$ is a low-rank parameter increment.
Having these motion-aware LoRAs, {\ours} is capable of synthesizing videos that are guided by both the textual description and the motion in the user-provided reference video.
\section{Experiments}
\label{sec:experiments}

\subsection{Experiment setup}
\label{sec:setup}

\paragraph{Dataset}
To evaluate {\ours}'s ability to transfer motion from a reference video to a new scene, we collect a dataset of 42 video-text pairs. These videos encompass a wide range of motion types, such as fast object movement, rotation, non-rigid motion, and camera movement. We also ensure that the scenes in the editing text prompts are distinct from the scene in the reference video while remaining compatible with its motion.

\paragraph{Implementation details}
 For a fair comparison, we use Zeroscope~\cite{zeroscope} as the base T2V diffusion model across all methods, given its ability to model complex motion and widespread usage in previous work~\cite{md,dmt,sma}. We fine-tune the model with LoRA~\cite{lora} for 400 steps at a learning rate of 0.0005. To extract motion features, we obtain attention maps $\mca$ and $\mtsa$ from down\_block.2, with weights $\wca$ and $\wtsa$ both set to 2000. These hyperparameters are chosen to balance control over camera framing and object movements. After extracting features from intermediate layers, we stop the forward pass to avoid unnecessary computation. For further implementation details, please refer to the supplementary material.

\paragraph{Baselines}
We compare our method against four recent approaches to motion customization, including two fine-tuning methods---VMC~\cite{vmc} and MotionDirector~\cite{md}---and two training-free methods---DMT~\cite{dmt} and MotionClone~\cite{mc}. Detailed descriptions of these methods are provided in \cref{sec:motion_customization}.

\begin{figure*}[t]
  \centering
  \includegraphics[width=1.0\linewidth]{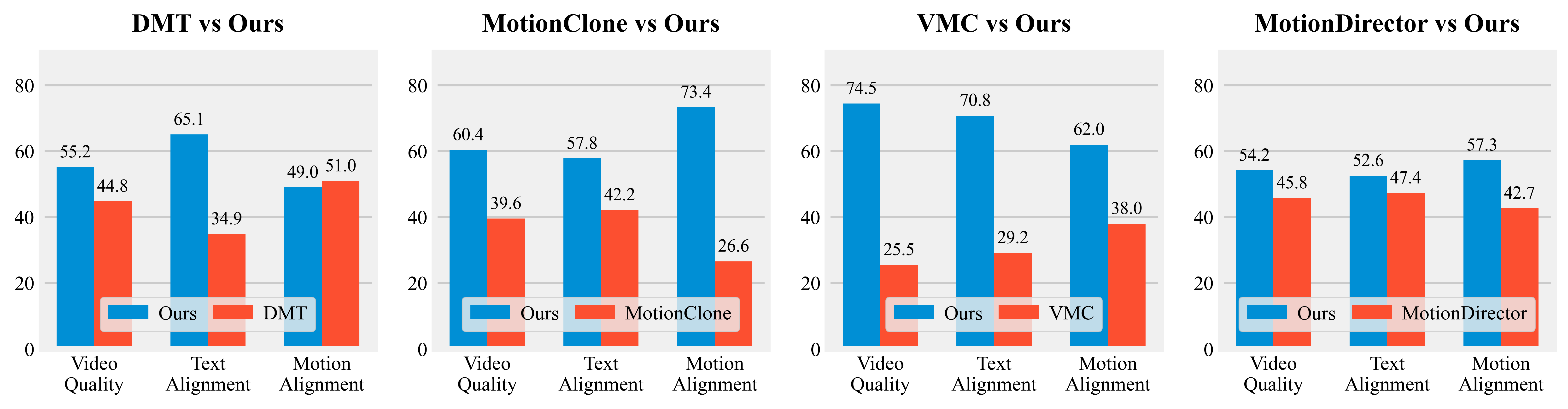}
  \caption{{\bf Human user study.} The results show that human raters prefer our method over existing approaches in terms of video quality, text alignment, and motion alignment.}
  \label{fig:human_study}
\end{figure*}
\begin{table*}[t]
  \centering
  \begin{tabular}{lcccc}
    \toprule
    Methods & CLIP-T ($\uparrow$) & ImageReward ($\uparrow$) & \makecell{Frame\\Consistency ($\uparrow$)} & \makecell{Motion\\Discrepancy ($\downarrow$)} \\
    \midrule
    DMT$^*$ & 29.19 & -0.0742 & 97.13 & \textbf{0.0284} \\
    MotionClone$^*$ & 29.69 & -0.1133	& 96.91 & 0.0503 \\
    VMC & 29.20 & -0.3292 & 96.89 & 0.0353 \\
    MotionDirector & \underline{30.31} & \underline{-0.0162} & \underline{97.19} & 0.0544 \\
    \textbf{Ours} & \textbf{30.43} & \textbf{0.2301} & \textbf{97.20} & \underline{0.0330} \\
    \bottomrule
  \end{tabular}
  \caption{{\bf Quantitative evaluation.} Our method outperforms baseline approaches in text alignment, frame consistency, and overall human preference as measured by ImageReward~\cite{ir}. Note that $^*$ denotes diffusion guidance-based methods.}
  \label{tab:quantitative}
\end{table*}
\begin{figure}[t]
  \centering
  \includegraphics[width=0.8\columnwidth]{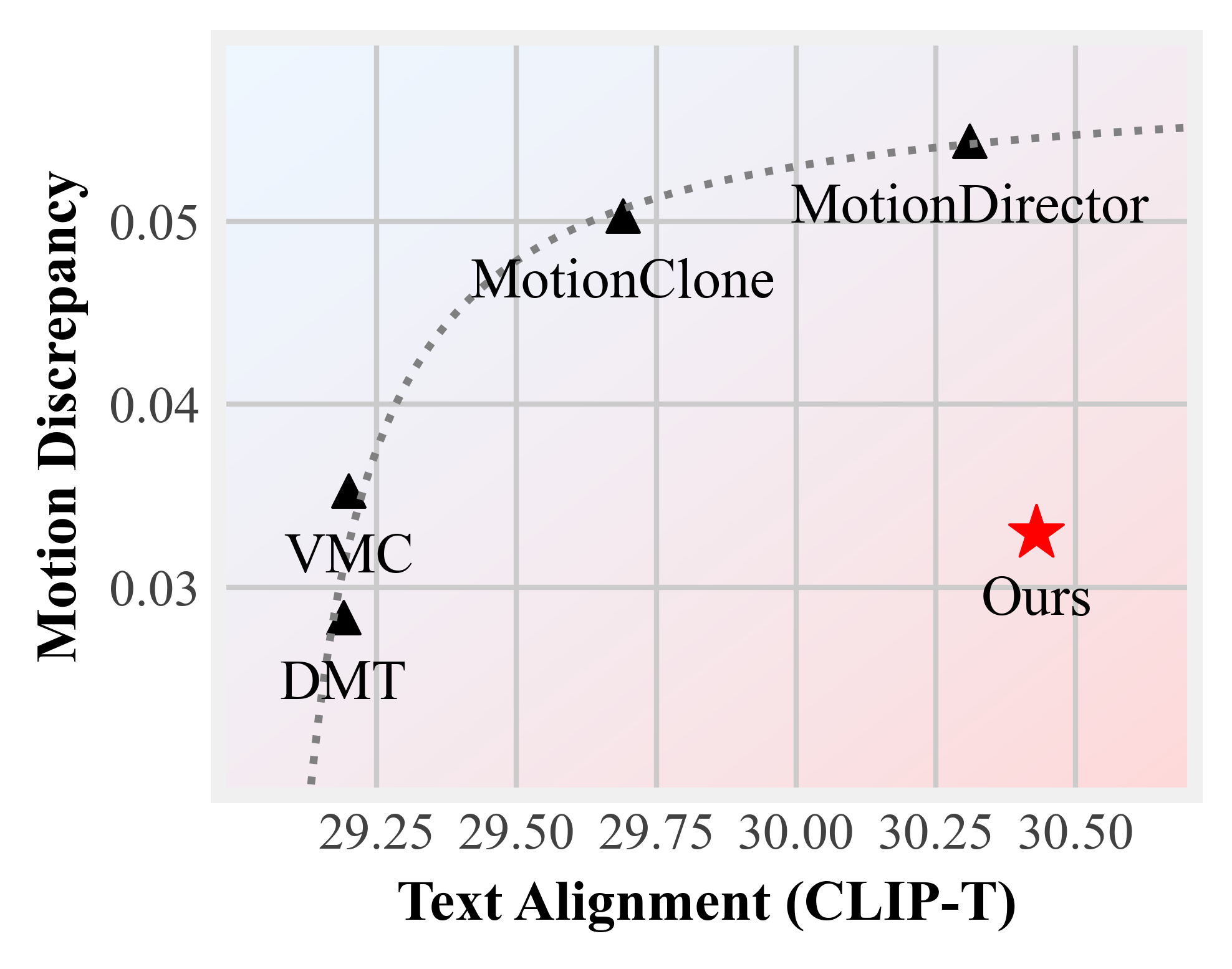}
  \caption{{\bf Illustration of the trade-off between text controllability and motion controllability.} The quantitative comparison shows that our framework is preferable due to better text alignment and lower motion discrepancy.}
  \label{fig:trade-off}
\end{figure}

\subsection{Evaluation metrics}
\label{sec:metrics}

We use four automatic metrics to evaluate the effectiveness of motion customization:
(1) {\bf CLIP-T:} To measure text alignment, we calculate the average CLIP~\cite{clip} cosine similarity between the text prompt and all output frames.
(2) {\bf Frame consistency:} We compute the average CLIP cosine similarity between each pair of consecutive frames to assess frame consistency.
(3) {\bf ImageReward:} We calculate the average ImageReward~\cite{ir} score for each frame, which evaluates both text alignment and image quality based on human preference.
(4) {\bf Motion discrepancy:} To quantify motion similarity between reference videos and generated videos, we leverage CoTracker3~\cite{cotracker3}, a state-of-the-art point tracker that densely tracks the motion trajectories of 2D points throughout a video. Specifically, we use CoTracker3 to generate $N$ 2D point trajectories for the reference video, denoted as $\hat{T}_0, \hat{T}_1,\cdots,\hat{T}_N\in \mathbb{R}^{F\times 2}$, and $N$ 2D point trajectories for the generated video, denoted as $T_0, T_1,\cdots,T_N\in \mathbb{R}^{F\times 2}$. To measure the similarity between these two sets of $F\times 2$ dimensional vectors, we use the Chamfer distance, a metric commonly used to assess the similarity between two sets of points in point cloud generation~\cite{pcg1,pcg2, pcg3, pcg4}. Accordingly, the {\it motion discrepancy} score is defined as:
\begin{equation}
C\left(\frac 1 N \sum_{i} \min_j \left\|T_i-\hat{T}_j\right\|^2 + \frac 1 N \sum_{j} \min_i \left\|T_i-\hat{T}_j\right\|^2\right)
\label{eq:motion_alignment},
\end{equation}
where $C=\frac 1 {2FHW}$ is a normalization constant.

\subsection{Main results}
\label{sec:quantitative}

\paragraph{Quantitative results}
The quantitative results are reported in \cref{tab:quantitative}. Our method outperforms all baseline approaches in metrics such as CLIP-T, frame consistency, and ImageReward, demonstrating its superiority in preserving the prior knowledge in the base model during fine-tuning.

We also visualize the trade-off between text controllability and motion controllability in \cref{fig:trade-off}. As shown, our method provides significantly better joint controllability of both text and motion than existing motion customization approaches.

\paragraph{Qualitative results}
In \cref{fig:qualitative}, we present qualitative comparisons with baseline approaches across various types of motion. In the first example, only our method successfully reproduces the fast displacement in the reference video, confirming the effectiveness of our motion feature extractor in capturing complex motion. In the second example, VMC and MotionClone misposition the object within the frame, whereas MotionDirector and DMT fail to generate realistic videos complying with the text prompt. In contrast, our method faithfully follows the text prompt and places the object correctly. In the third and forth examples, our method also exhibits superior visual and motion quality.

These results conclude that our method preserves \emph{the most} pre-trained knowledge during fine-tuning, while providing \emph{the strongest} controllability for complex motion. For more results, please refer to \cref{fig:teaser} and the appendix.

\section{Ablation study}
\label{sec:ablation}

\begin{figure}[t]
  \centering
  \includegraphics[width=1.0\columnwidth]{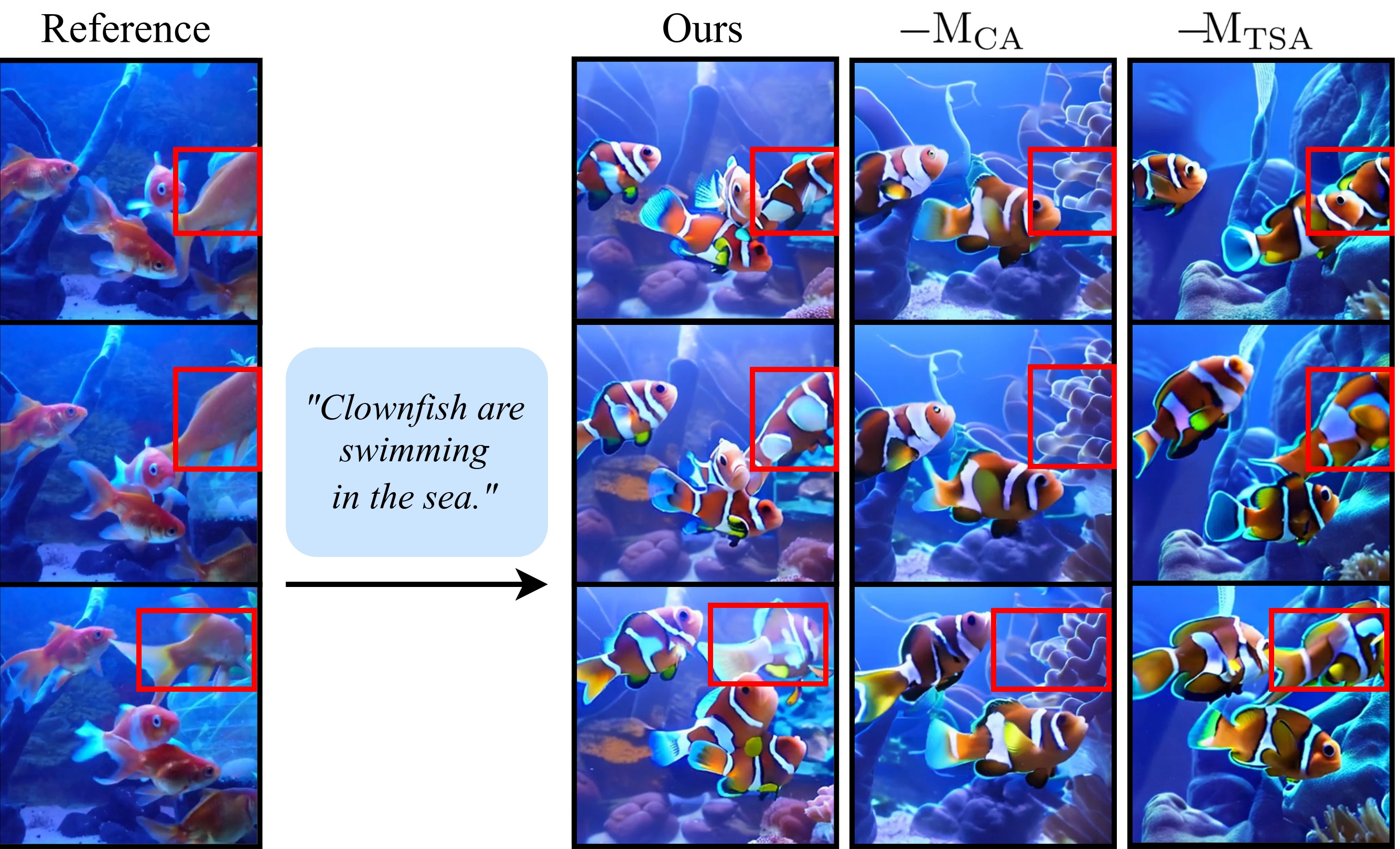}
  \caption{\textbf{Qualitative results for ablation study.} Without utilizing cross-attention maps $\mca$ in motion features, the model fails to capture all the fish in the video, whereas in the absence of temporal self-attention maps $\mtsa$, the model struggles to accurately replicate the fine-grained motion details. In contrast, our method successfully preserves both the scene composition and the inter-frame dynamics of the reference video.}
  \label{fig:ablation}
\end{figure}

We conduct an ablation study to examine the impact of incorporating $\mca$ and $\mtsa$ in motion features. As illustrated in \cref{fig:ablation}, without cross-attention maps $\mca$, the model struggles to correctly position all the element of the scene. Meanwhile, removing temporal self-attention maps $\mtsa$ reduces the precision of fine-grained dynamics. The quantitative results in \cref{tab:ablation_quant} further validate the importance of both attention maps in controlling motion. These results confirm that both the \emph{camera framing}, informed by $\mca$, and \emph{inter-frame dynamics}, informed by $\mtsa$, are essential for capturing overall motion.

\subsection{Human user study}

For a more accurate evaluation, we conduct a user study comparing our method with existing approaches based on human preferences. Following previous work~\cite{md,dmt}, we adopt the Two-alternative Forced Choice (2AFC) protocol. In the survey, the participants are presented with one video generated by our method and another video generated by a baseline approach. They are asked to compare the videos across three key aspects of motion customization:
(1) {\bf Video quality:}  the degree to which the output video appears realistic and visually appealing,
(2) {\bf Text alignment:} how well the output video matches the text prompt, and
(3) {\bf Motion alignment:} the similarity in motion between the output video and the reference video. Ultimately, we collected 192 human evaluations per baseline and metric, totaling 2,304 human evaluations. These responses were gathered from 24 participants recruited via the Prolific platform.

As shown in \cref{fig:human_study}, human users prefer our method over existing approaches in all aspects. These results further confirm the superiority of our method.

\begin{table}[t]
  \centering
\begin{tabular}{ccccc}
  \toprule
  &CLIP-T ($\uparrow) $ &ImageReward ($\uparrow$) & \makecell{Motion\\Discrep. ($\downarrow$)} \\ \midrule
  $-$CA&30.08&0.1252&0.0360 \\
  $-$TSA&30.67&0.4650&0.0693 \\
  Ours&30.43&0.2301&0.0330 \\
  \bottomrule
  \end{tabular}
  \caption{{\bf Ablation study.} Our method, which utilizes both $\mca$ and $\mtsa$, achieves the lowest motion discrepancy score.}
  \label{tab:ablation_quant}
\end{table}
\section{Conclusion}
\label{sec:conclusion}

We presented \ours, a feature-level fine-tuning framework for motion customization. {\ours} transforms the \emph{pixel-level} DDPM objective into the \emph{motion feature matching} objective, aiming to learn the target motion at the \emph{feature level}. To extract motion features, {\ours} leverages the pre-trained T2V diffusion model as a deep feature extractor and identify valuable motion cues from two attention mechanisms within the model, representing both object movements and camera framing in videos. In the experiments, {\ours} demonstrated superior joint controllability of text and motion to prior approaches. These results suggest that {\ours} enhances control over scene staging in AI-generated videos, benefiting  real-world applications in computer-generated imagery (CGI). For a discussion of {\ours}'s limitations, please refer to the supplementary material.
{
    \small
    \bibliographystyle{ieeenat_fullname}
    \bibliography{main}
}
\clearpage
\appendix
\setcounter{page}{1}
\maketitlesupplementary

\section{Extended derivations}
\label{sec:derivation}

Below is the derivation of Eq. (2). We apply the generalized formula in DDIM~\cite{ddim} to compute the less noisy video at timestep $t-1$ (denoted as $v_t$), using the noisy video $z_t$ at timestep $t$ along with the predicted noise $\epsilon$:
\begin{align}
    v_{t}(\epsilon,z_t)=
&\sqrt{\bar\alpha_{t-1}} \underbrace{\left(\frac{z_t-\sqrt{1-\bar\alpha_t}\epsilon}{\sqrt{\bar\alpha_t}}\right)}_{\text{``predicted } z_0 \text{''}}\notag\\
&+\underbrace{\sqrt{1-\bar\alpha_{t-1}-\sigma_t^2}\cdot\epsilon}_{\text{``direction pointing to } z_t \text{''}}\notag\\
&+\underbrace{\sigma_t\epsilon_t}_{\text{random noise}}
  \label{eq:derivation}
\end{align}
where $\bar\alpha_t:=\prod_{s=1}^t\alpha_s$ are variance-scaling coefficients~\cite{ddpm}, $\epsilon_t\sim\mathcal N(\bf{0},\bf{I})$ is Gaussian noise, and $\sigma$ is a hyperparamter controlling the stochasticity of the sampling process.

We observe that reducing randomness (\ie using a lower value of $\sigma_t$) improves feature extraction. Thus, following DDIM, we set $\sigma_t=0$. This simplifies the equation to:
\begin{align}
    v_{t}=
\sqrt{\bar\alpha_{t-1}} \left(\frac{z_t-\sqrt{1-\bar\alpha_t}\epsilon}{\sqrt{\bar\alpha_t}}\right)
+\sqrt{1-\bar\alpha_{t-1}}\cdot\epsilon
\end{align}
which can be further simplified as:
\begin{align}
    v_{t}=
&\frac 1 {\sqrt{\alpha_{t}}}z_t+\left(-\frac {\sqrt{1-\bar\alpha_t}} {\sqrt{\alpha_{t}}}+\sqrt{1-\bar\alpha_{t-1}}\right)\epsilon
\end{align}
Next, the DDPM objective can be reformulated to compare the previous noised videos $z_{t-1}$:
\begin{align}
    L=&\E\left[w_t\left\|
\epsilon-\epsilon_{\theta}(z_t, t, c)
\right\|^2\right]\\
=&\E\left[w'_t\left\|
v_t(z_t,\epsilon)-v_t(z_t,\epsilon_{\theta}(z_t, t, c))
\right\|^2\right]
\end{align}
where:
\begin{align}
w'_t=\left(-\frac {\sqrt{1-\bar\alpha_t}} {\sqrt{\alpha_{t}}}+\sqrt{1-\bar\alpha_{t-1}}\right)^{-1}w_t
\end{align}
The time-dependent weight $w_t$ is commonly set to 1. However, we employ a different weighting, where $w'_t$ is 1 for the first 500 steps and to 0 for the last 500 steps. This weighting approach prioritizes the early stages, which are crucial for deciding video motion.

\section{Limitations}
\label{sec:limitations}

One limitation of {\ours} is that it requires a feature extractor to compute the objective, which introduces additional latency and results in longer training time (15 minutes) compared to pixel-level fine-tuning approaches~\cite{md,vmc} (8 minuets) on an NVIDIA GeForce RTX 4090. Furthermore, since {\ours} relies on pre-trained T2V diffusion models, it struggles to synthesize videos that fall outside the generative prior of these models. However, we believe that this challenge can be mitigated as more advanced T2V diffusion models are developed in the future.

Like other existing approaches, another limitation of {\ours} lies in its reliance on DDIM-inverted noise (See \cref{sec:noise} for details), which introduces a potential risk of content leakage from the reference video. As this issue is common among most existing approaches, addressing it will be an important direction for future research.

\section{Analysis of motion features}
\label{sec:supp_retrieval}

We conduct a simple retrieval experiment to verify that our motion feature extractor is capturing motion information from noisy videos. From the SVW dataset~\cite{svw}, we draw 139 javelin video clips with diverse motion trajectories and camera movements and randomly trim each clip to 16 frames. We obtain their motion features by adding noise to each video $z$ and feeding them into our motion feature extractor as follows:
\begin{equation}
\mathcal \mfe(\sqrt{\bar{\alpha}_t}z+\sqrt{1-\bar{\alpha}_t}\epsilon)
\label{eq:retrieval},
\end{equation}
where $\mfe$ denotes our motion feature extractor, and the time step $t$ is set to $500$ for this experiment. After getting the motion features of all videos, we randomly select a query video and retrieve the most similar video from the dataset based on these motion features.

\begin{figure}[t]
  \centering
  \includegraphics[width=1.0\columnwidth]{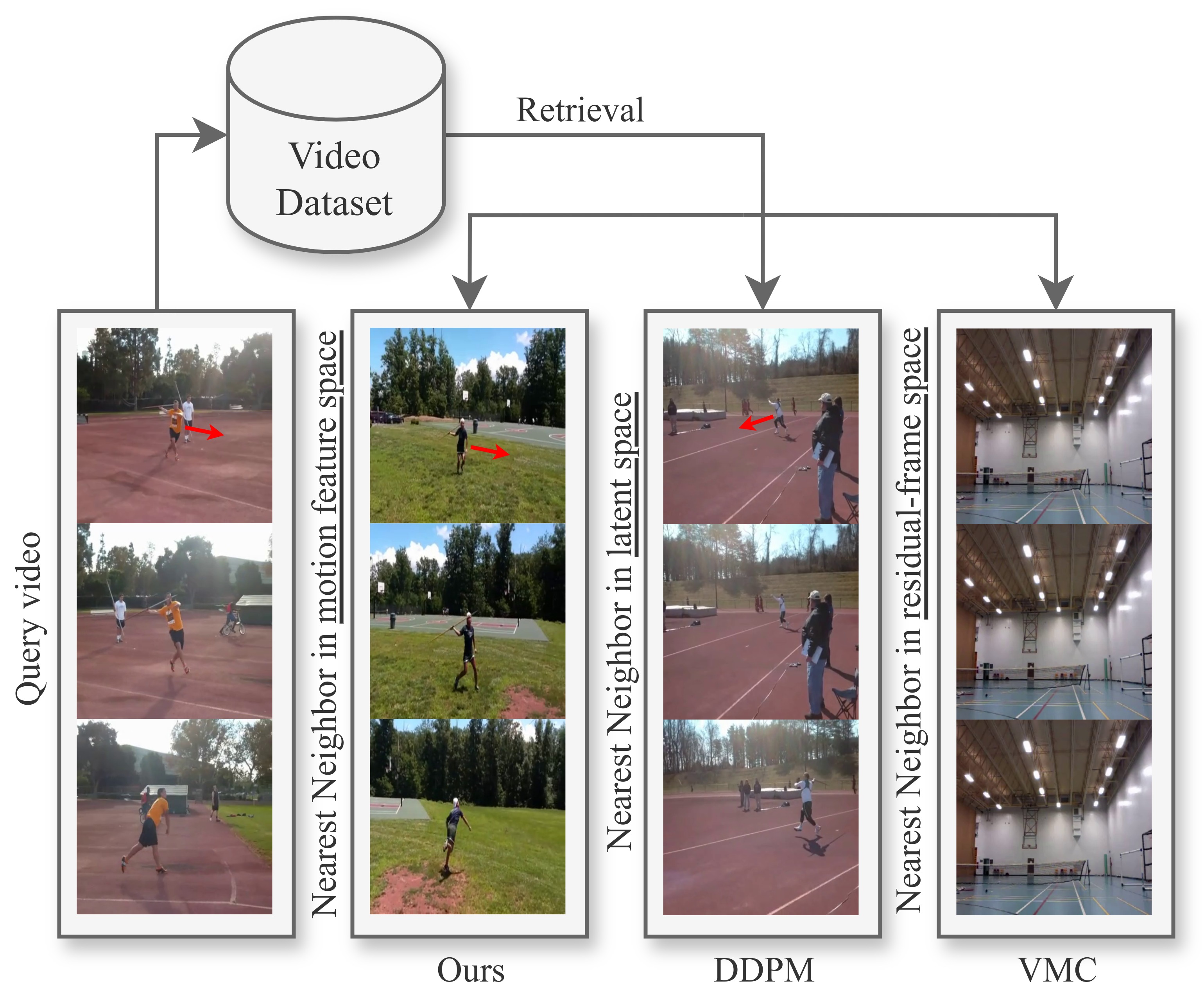}
  \caption{{\bf Motion Retrieval.} Compared to DDPM~\cite{ddpm} (using latent values) and VMC~\cite{vmc} (using frame differences), using the proposed motion features to perform motion retrieval shows preferable results. Note that the nearest neighbor in the motion feature space is retrieved by matching the motion features of the query video with those of the video dataset.}
  \label{fig:retrieval}
\end{figure}

As shown in \cref{fig:retrieval}, the video with the most similar motion features shares the same motion despite having different appearances. In contrast, the video that is most similar in latent space has a nearly identical appearance but opposite motion, while the video with the most similar residual frames contain unrelated motion.

To compute the retrieval accuracy statistically, we label the videos with the top 10\% smallest motion discrepancy values with the query video as positive samples and the rest 90\% of the videos as negative samples. Next, we compute the average precisions (AP) for each retrieval methods to assess their retrieval accuracy. As presented in \cref{tab:retrieval_quant}, our motion features yield the highest accuracy, indicating that they have the strongest correlation with actual motion. These results verify that our motion features capture rich motion information, rather than irrelevant details about visual appearance.

\begin{table}[h]
  \centering
\begin{tabular}{ccccc}
  \toprule
  &\textbf{Ours} & DDPM & VMC & Random \\ \midrule
  AP & \textbf{32.78\%} & 8.20\% & 8.85\% &10.71\% \\
  \bottomrule
\end{tabular}

  \caption{{\bf Retrieval accuracy.} Using our motion features to extract videos with similar motion yields the highest average precision (AP) than directly using latent videos (DDPM~\cite{ddpm}) or their residual frames (VMC~\cite{vmc}).}
  \label{tab:retrieval_quant}
\end{table}

\section{Additional qualitative results}
\label{sec:supp_qualitative}

\begin{figure*}
  \centering
  \includegraphics[width=1.0\linewidth]{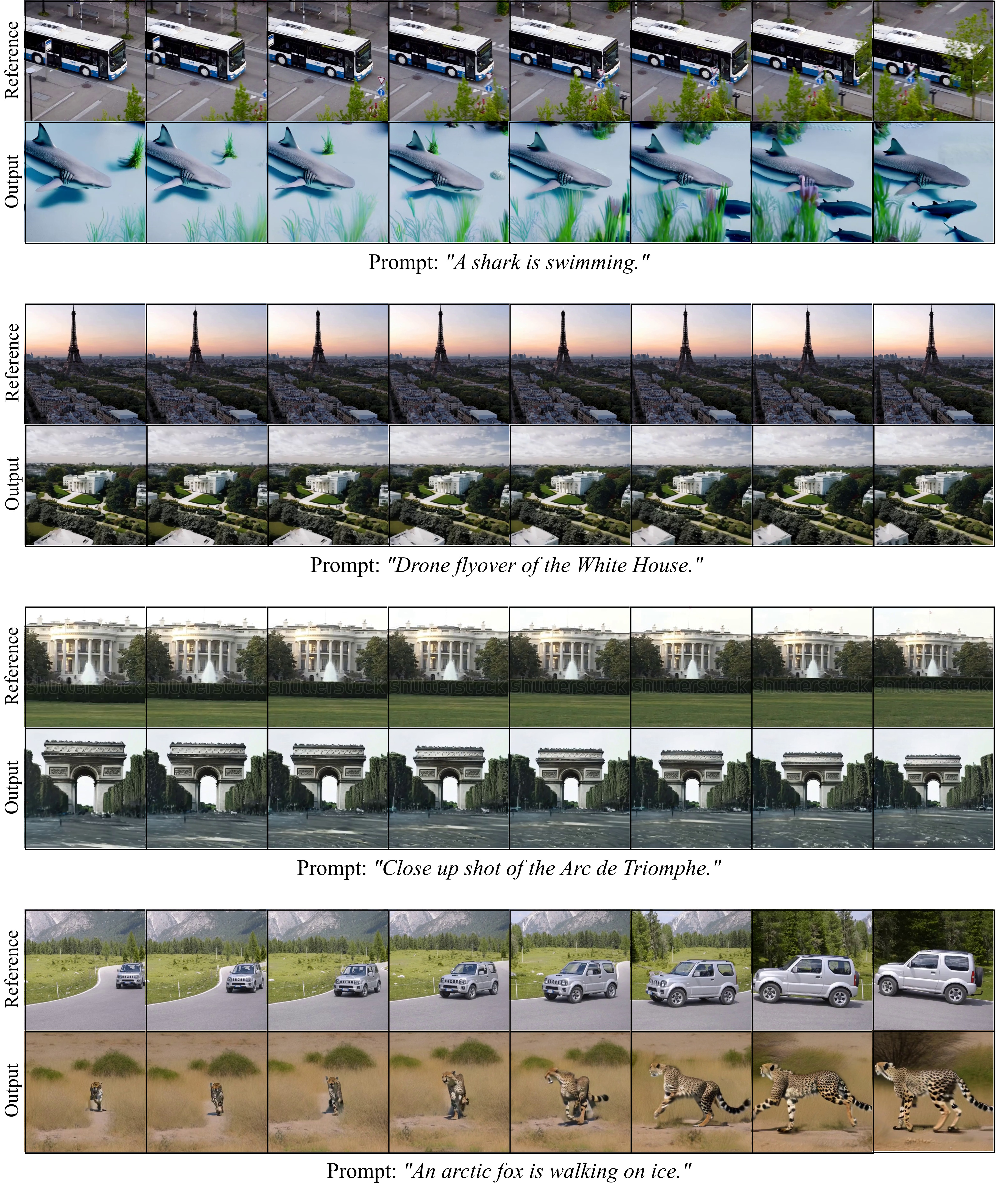}
  \caption{{\bf Additional qualitative results.} The results demonstrate {\ours}'s capability to transfer both object movements and camera movements to new scenes.}
  \label{fig:supp_qualitative}
\end{figure*}

\begin{figure*}
  \centering
  \includegraphics[width=1.0\linewidth]{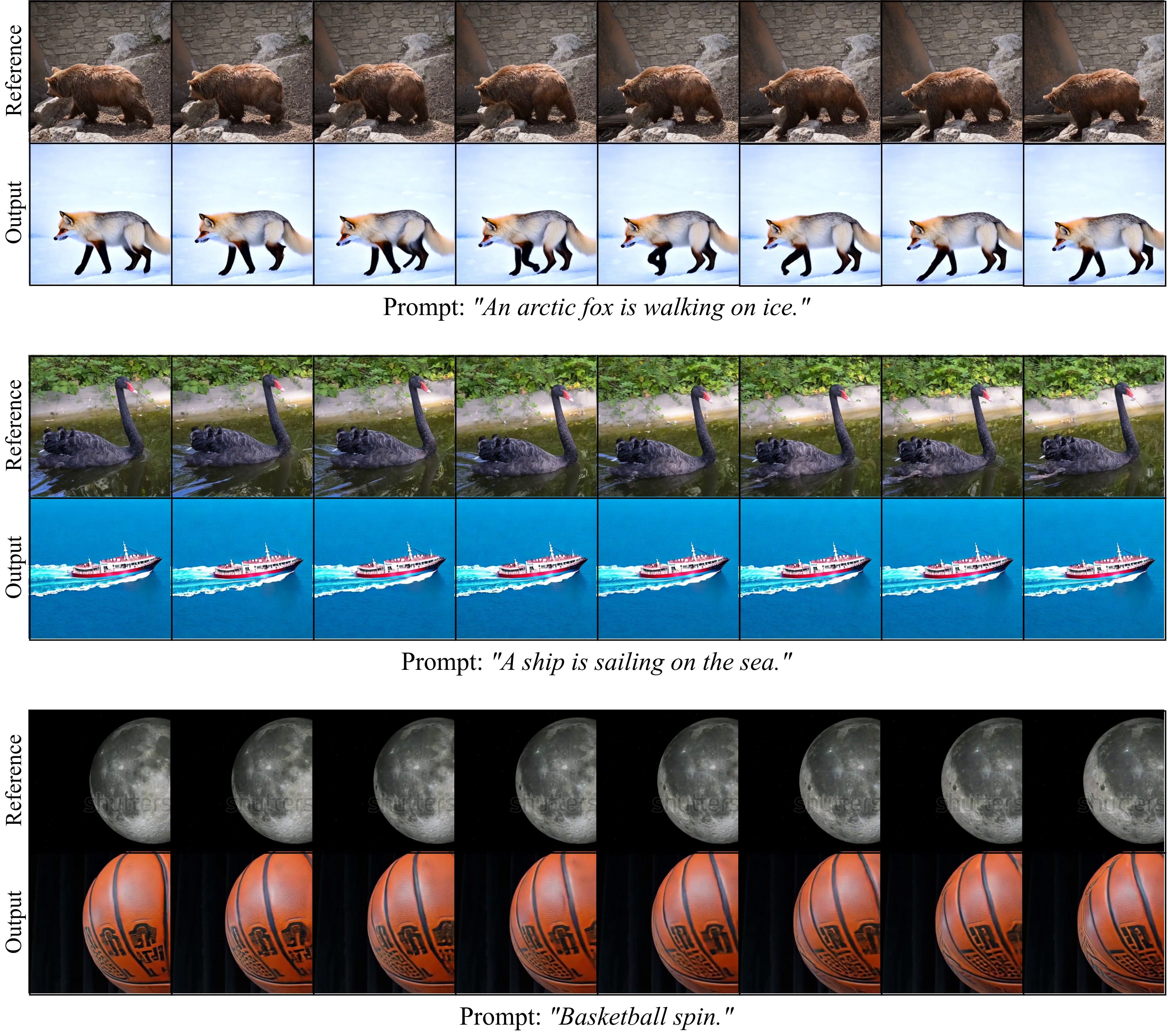}
  \caption{{\bf More samples generated using CogVideoX~\cite{cogvideo} as the base model.} The results demonstrate the generality of {\ours}. Even with T2V diffusion models that employ full attentions, we can still extract cues for objects movement from attention weights computed between frames and cues for camera framing from attention weights computed between words and patch tokens.}
  \label{fig:cogvideo}
\end{figure*}

\begin{figure*}
  \centering
  \includegraphics[width=1.0\linewidth]{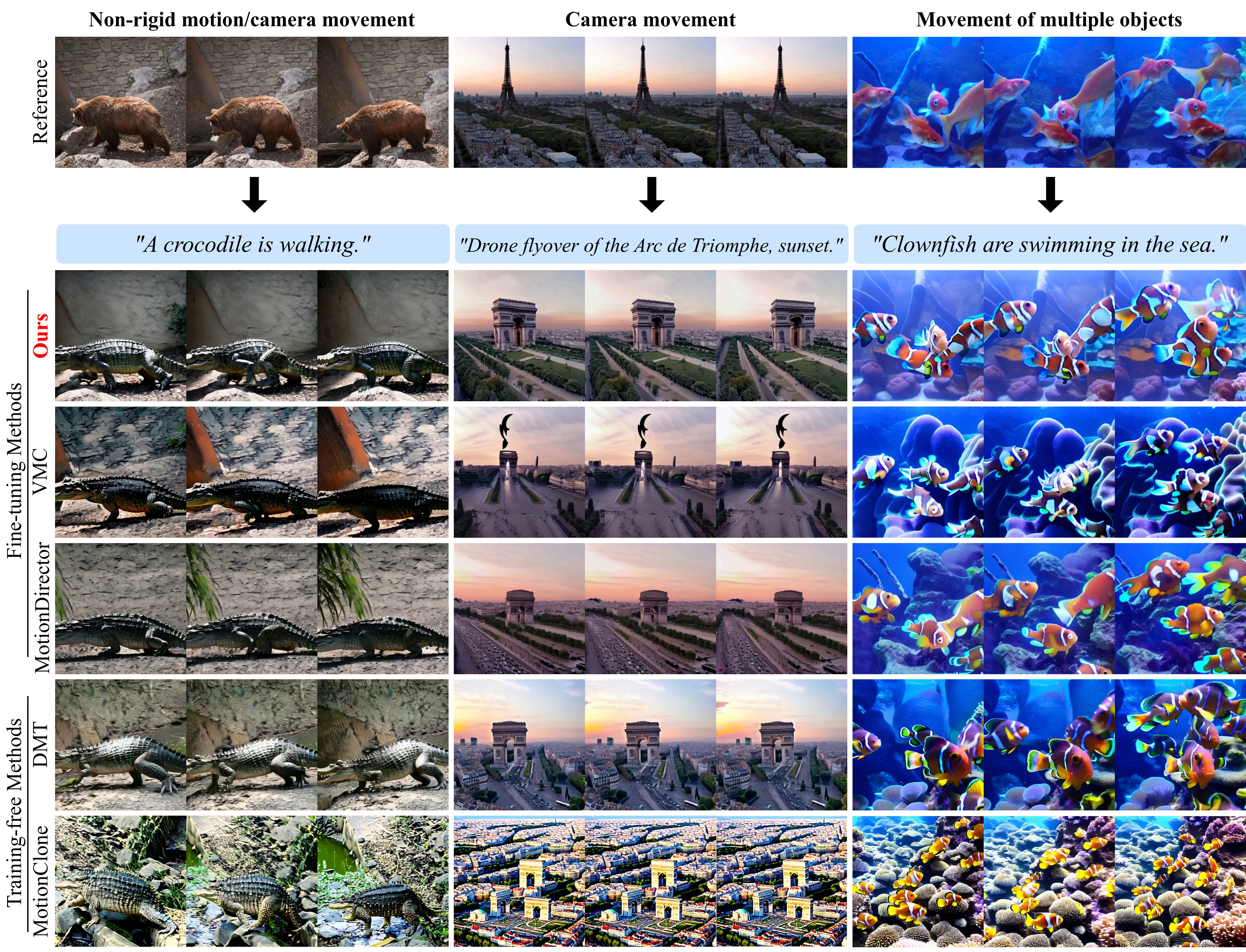}
  \caption{{\bf Additional qualitative comparisons.} The results demonstrate {\ours}'s superiority over existing motion customization methods in terms of video quality, text alignment, and motion alignment.}
  \label{fig:supp_comparisons}
\end{figure*}

We present additional qualitative comparisons in \cref{fig:supp_comparisons}, detailed qualitative results in \cref{fig:supp_qualitative}, and further samples generated using CogVideoX~\cite{cogvideo} as the base model in \cref{fig:cogvideo}.

\section{Must motion be learned at feature level?}
\label{sec:why}

Analyzing video motion requires the ability to identify (1) scene composition and (2) the patterns of changes across frames (\ie zooming, rotation, and displacement). Both of them are high-level concepts. The high-level nature of motion is also evident in optical flow estimation, a longstanding focus of research in video motion analysis. Early efforts in this domain primarily relies on rule-based algorithms that use handcrafted rules to model motion~\cite{algo1, algo2, algo3, algo4}. However, such methods often struggle with complex motion, such as large displacements, non-rigid movements, and motion in low-texture regions, all due to their lack of high-level understanding of videos.

With advances in machine learning, recent studies on optical flow estimation have shifted towards data-driven methods that learn motion patterns from large datasets~\cite{deep1, deep2, deep3, deep4, deep5}. These approaches have significantly improved motion estimation by leveraging deep neural networks to understand motion at the feature level, highlighting the importance of a high-level understanding of motion. 

In the context of motion customization, given that motion is inherently a high-level concept, pixel-level objectives, such as frame-difference matching~\cite{md,vmc,sma}, are insufficient for capturing motion. These objectives often fail to capture complex motion, facing the same challenge as early research on optical flow estimation. In contrast, our method precisely extracts motion information with the assistance of a deep neural network. By leveraging a large pre-trained model, our method can understand at a high level and captures key information such as scene composition and patterns of changes.

\section{Implementation details}
\label{sec:rationale}

\paragraph{Training}
\label{sec:supp_training}
To fine-tune the diffusion model, we add LoRAs to all self-attention and feed forward layers, and set the rank to 32. Since motion is mainly determined in early stages~\cite{mc,dmt}, we set the time-dependent weights $w'_t$ in the objective function to 1 for the first 500 timesteps and 0 for the last 500 timesteps. The LoRA~\cite{lora} are optimized for 400 steps at a learning rate of 0.0005, which takes approximately 15 minutes on an NVIDIA GeForce RTX 4090. All videos in the experiments consist of 16 frames at 8 fps and are generated at a resolution of $384\times 384$.
  
\paragraph{Feature extraction}
We extract cross-attention maps and temporal-self attention maps from down\_block.2 at a $12\times 12$ resolution. Both $\mca$ and $\mtsa$ represent the average of all extracted attention maps across heads and layers, which we omit in all equations for conciseness.

\paragraph{Initial noise}
\label{sec:noise}
Following previous work on motion customization~\cite{md, vmc, sma, dmt}, we utilize DDIM inversion to obtain the initial noise $z_T$ for better motion alignment. In our work, the initial noise $z_T$ is computed as in MotionDirector's implementation:
\begin{equation}
    z_T=\sqrt{\beta}\epsilon_{\rm inv}+\sqrt{1-\beta}\epsilon
  \label{eq:ddpm_in}
\end{equation}
where $\epsilon\sim\mathcal N(\bf{0},\bf{I})$ is Gaussian noise, and $\epsilon_{\rm inv}$ represents the inverted noise of the reference video, derived via DDIM inversion~\cite{ddim}. The square root terms in the equation ensure that the variance of $z_T$ remains consistent across all values of $\beta$. In quantitative experiments and human user study, we set a fix value of $\beta=0.3$. In other experiments, $\beta$ varies between the range of $0.0$ to $0.3$.

\section{Evaluation details}

\paragraph{Dataset}
We collect a dataset of 42 video-text pairs, including 14 unique reference videos from DAVIS~\cite{davis} and LOVEU-TGVE~\cite{loveu}, many of which are also used in prior work. For each reference video, we provide exactly 3 target text prompts that describe scenes distinct from the original one and ensure that they are compatible with the motion in the reference video.

\paragraph{Quantitative evaluation}
To evaluate each method, we generate 5 videos per video-text pair, and calculate the average scores across all generated videos.

\begin{figure*}
  \centering
  \includegraphics[width=0.7\linewidth]{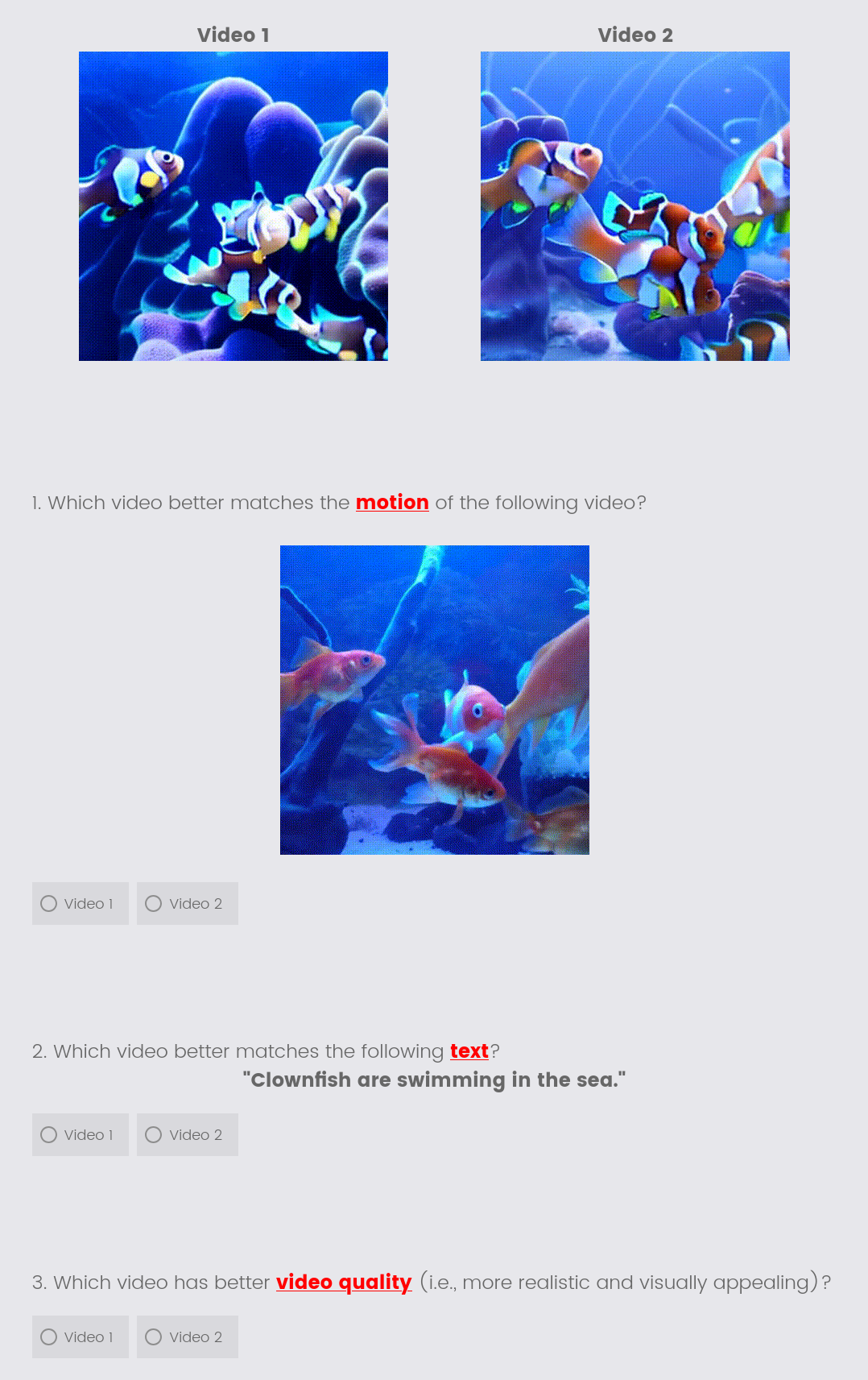}
  \caption{{\bf User interface of an evaluation task.} Each task includes three questions, each assessing a key aspect of motion customization.}
  \label{fig:ui}
\end{figure*}

\paragraph{Human user study}
In the human user study, we employ the same set of videos generated in the quantitative experiments. Each survey consists of 32 tasks. In each task, the survey respondents are presented with a video-text pair, a video generated by our method, and a video generated by one of the four competing methods (\cref{fig:ui}). The video-text pair and videos for each task are randomly selected on the fly, resulting in a total of $4\times 42\times 5\times 5=4200$ different tasks. To assess motion alignment, text alignment, and video quality, the participants are asked three questions: "Which video better matches the motion of the following video?", "Which video better matches the following text?", and "Which video has better video quality (i.e., more realistic and visually appealing)?". To ensure a fair comparison, the order of the choices is randomized.

\clearpage

\clearpage

\clearpage

\clearpage

\end{document}